\documentclass[10pt,onecolumn]{article}

\usepackage{times}
\usepackage[dvips]{graphicx}
\usepackage[hang,scriptsize]{subfigure}
\usepackage{latexsym}

\usepackage{amsmath} 
\usepackage{amssymb} 
\usepackage{graphics} 

\makeatletter
\def\@maketitle{%
  \vbox to 6.5cm{%
    \hsize\textwidth
    \linewidth\hsize
    \vspace{1.5cm}
    \centering
    {\bfseries\LARGE \@title \par}
    \vspace{24pt}
    {\fontsize{11pt}{13pt}\selectfont \begin{tabular}[t]{c}\@author \end{tabular}\par}
    \vfill} 
}
\renewcommand\section{\@startsection{section}{1}{\z@}%
                       {-12\p@ \@plus -4\p@ \@minus -4\p@}%
                       {6\p@ \@plus 4\p@ \@minus 4\p@}%
                       {\normalfont\large\bfseries
                        \rightskip=\z@ \@plus 8em\pretolerance=10000 }}
\renewcommand\subsection{\@startsection{subsection}{2}{\z@}%
                       {-12\p@ \@plus -4\p@ \@minus -4\p@}%
                       {6\p@ \@plus 4\p@ \@minus 4\p@}%
                       {\normalfont\fontsize{11pt}{13pt}\selectfont\bfseries
                        \rightskip=\z@ \@plus 8em\pretolerance=10000 }}
\renewcommand\subsubsection{\@startsection{subsubsection}{3}{\z@}%
                       {-12\p@ \@plus -4\p@ \@minus -4\p@}%
                       {6\p@ \@plus 4\p@ \@minus 4\p@}%
                       {\normalfont\normalsize\itshape}}
\renewcommand\paragraph{\@startsection{paragraph}{4}{\z@}%
                       {-12\p@ \@plus -4\p@ \@minus -4\p@}%
                       {-0.5em \@plus -0.22em \@minus -0.1em}%
                       {\normalfont\normalsize\itshape}}
\makeatother

\renewenvironment{abstract}%
  {\small
    \list{}{\labelwidth0pt
      \leftmargin0pt \rightmargin\leftmargin
      \listparindent\parindent \itemindent0pt
      \parsep0pt
      }%
    \item[\hskip\labelsep\bfseries\abstractname\enspace --] \itshape}{\endlist}

\newcommand{\keywordsname}{Keywords}
\newenvironment{keywords}%
  {\small
    \list{}{\labelwidth0pt
      \leftmargin0pt \rightmargin\leftmargin
      \listparindent\parindent \itemindent0pt
      \parsep0pt
      }%

    \item[\hskip\labelsep\bfseries\keywordsname:]}{\endlist}

\begin{document}

\title{A Simple Proportional Conflict Redistribution Rule}

\author{\begin{tabular}{c@{\extracolsep{4em}}c@{\extracolsep{1em}}c}
{\bf Florentin Smarandache} & {\bf Jean Dezert}\\
Dept. of Mathematics &  ONERA/DTIM/IED \\
Univ. of New Mexico  & 29 Av. de la  Division Leclerc \\
Gallup, NM 8730 & 92320 Ch\^{a}tillon \\
U.S.A.  & France \\
{\tt smarand@unm.edu} & {\tt Jean.Dezert@onera.fr}
\end{tabular}}

\date{}
\maketitle
\pagestyle{plain}

\begin{abstract}
One proposes a first alternative rule of combination to WAO (Weighted Average Operator) proposed recently by Josang, Daniel and Vannoorenberghe, called Proportional Conflict Redistribution rule (denoted PCR1). PCR1 and WAO are particular cases of WO (the Weighted Operator) because the conflicting mass is redistributed with respect to some weighting factors.  In this first PCR rule, the proportionalization is done for each non-empty set with respect to the non-zero sum of its corresponding mass matrix - instead of its mass column average as in WAO, but the results are the same as Ph. Smets has pointed out. Also, we extend WAO (which herein gives no solution) for the degenerate case when all column sums of all non-empty sets are zero, and then the conflicting mass is transferred to the non-empty disjunctive form of all non-empty sets together; but if this disjunctive form happens to be empty, then one considers an open world (i.e. the frame of discernment might contain new hypotheses) and thus all conflicting mass is transferred to the empty set. In addition to WAO, we propose a general formula for PCR1 (WAO for non-degenerate cases).  Several numerical examples and comparisons with other rules for combination of evidence published in literature are presented too. Another distinction between these alternative rules is that WAO is defined on the power set, while PCR1 is on the hyper-power set (Dedekind's lattice). A nice feature of PCR1, is that it  works not only on non-degenerate cases but also on degenerate cases as well appearing in dynamic fusion, while WAO gives the sum of masses in this cases less than 1 (WAO does not work in these cases). Meanwhile we show that PCR1 and WAO do not preserve unfortunately the neutrality property of the vacuous belief assignment though the fusion process. This severe drawback can however be easily circumvented by new PCR rules presented in a companion paper. 
\end{abstract}


\section{Introduction}

Due to the fact that Dempster's rule is not mathematically defined for conflict 1 or gives counter-intuitive results for high conflict (see Zadeh's example \cite{Zadeh_1984}, Dezert-Smarandache-Khoshnevisan's examples \cite{DSmTBook_2004a}), we looked for another rule, similar to Dempster's, easy to implement due to its simple formula, and working in any case no matter the conflict. We present this PCR1 rule of combination, which is an alternative of WAO for non-degenerate cases, in many examples comparing it with other existing rules mainly: Smets', Yager's, Dubois-Prade's, DSm hybride rule, Murphy's, and of course Dempster's.  PCR1 rule is commutative, but not associative nor Markovian (it is however quasi-associative and quasi-Markovian). More versions of PCR rules are proposed in a companion paper \cite{DSmTBook_2004b} to overcome the limitations of PCR1 presented in the sequel.

\section{Existing rules for combining evidence}

We briefly present here the main rules proposed in the literature for combining/aggregating several independent and equi-reliable sources of evidence expressing their belief on a given finite set of exhaustive and exclusive hypotheses (Shafer's model). We assume the reader familiar with the Dempster-Shafer theory of evidence \cite{Shafer_1976} and the recent theory of plausible and paradoxical reasoning (DSmT) \cite{DSmTBook_2004a}. A detailed presentation of these rules can be found in \cite{DSmTBook_2004a} and \cite{Sentz_2002}. In the sequel, we consider the Shafer's model as the valid model for the fusion problem under consideration, unless specified.\\

Let $\Theta=\{\theta_1,\theta_2,\ldots,\theta_n\}$ be the {\it{frame of discernment}} of the fusion problem under consideration having $n$ {\it{exhaustive}} and {\it{exclusive elementary}} hypotheses $\theta_i$. 
The set of all subsets of $\Theta$ is called the {\it{power set}} of $\Theta$ and is denoted $2^\Theta$. Within Shafer's model, a {\it{basic belief assignment}} (bba) $m (.): 2^\Theta \rightarrow  [0, 1]$  associated to a given body of evidence $\mathcal{B}$ is defined by \cite{Shafer_1976}
\begin{equation}
m(\emptyset)=0  \qquad \text{and}\qquad     \sum_{X\in 2^\Theta} m(X) = 1    
\label{eq:bba}                        
\end{equation}
\noindent
The belief (credibility) and plausibility functions of $X\subseteq\Theta$ are defined as
\begin{equation}
\text{Bel}(X) = \sum_{Y\in 2^\Theta, Y\subseteq X} m(Y)
\label{Belg}
\end{equation}
\begin{equation}
\text{Pl}(X) = \sum_{Y\in 2^\Theta, Y\cap X\neq\emptyset} m(Y)=1- \text{Bel}(\bar{X})
\label{Plg}
\end{equation}
\noindent 
where $\bar{X}$ denotes the complement of $X$ in $\Theta$. \\

The belief functions $m(.)$, $\text{Bel}(.)$ and $\text{Pl}(.)$ are in one-to-one correspondence. The set of elements $X\in 2^\Theta$ having a positive basic belief assignment  is called the {\it{core/kernel}} of the source of evidence under consideration.\\

The main problem is now how to combine several belief assignments provided by a set of independent sources of evidence. This problem is fundamental to pool correctly uncertain and imprecise information and help the decision-making. Unfortunately, no clear/unique and satisfactory answer to this problem exists since there is potentially an infinite number of possible rules of combination \cite{Inagaki_1991, Lefevre_2002, Sentz_2002}. Our contribution here is to propose an alternative to existing rules which is very easy to implement and have a legitimate behavior (not necessary the optimal one - if such optimality exists ...) for practical applications.

\subsection{The Dempster's rule }

The Dempster's rule of combination is the most widely used rule of combination so far in many expert systems based on belief functions since historically it was proposed in the seminal book of Shafer in \cite{Shafer_1976}. This rule, although presenting interesting advantages (mainly the commutativity and associativity properties) fails however to provide coherent results due to the normalization procedure it involves. Discussions on the justification of the Dempster's rule and its well-known limitations can be found by example in \cite{Zadeh_1979,Zadeh_1984,Zadeh_1985,Voorbraak_1991}. The Dempster's rule is defined as follows: let $\text{Bel}_1(.)$ and $\text{Bel}_2(.)$ be two belief functions provided by two independent equally reliable sources of evidence $\mathcal{B}_1$ and $\mathcal{B}_2$ over the same frame $\Theta$ with corresponding belief assignments $m_1(.)$ and $m_2(.)$. Then the combined global belief function denoted $\text{Bel}(.)= \text{Bel}_1(.)\oplus \text{Bel}_2(.)$ is obtained by combining $m_1(.)$ and $m_2(.)$ according to $m(\emptyset)=0$ and $\forall (X\neq\emptyset) \in 2^\Theta$ by

\begin{equation}
m (X) = \frac{\displaystyle\sum_{\substack{X_1,X_2\in 2^\Theta \\ X_1\cap X_2=X}}m_{1}(X_1)m_{2}(X_2)}
{\displaystyle 1-\sum_{\substack{X_1,X_2\in 2^\Theta\\ X_1\cap X_2=\emptyset}} m_{1}(X_1) m_{2}(X_2)}
\label{eq:DSR}
 \end{equation}

$m(.)$ is a proper basic belief assignment if and only if the denominator in equation \eqref{eq:DSR} is non-zero. The  {\it{degree of conflict}} between the sources $\mathcal{B}_1$ and $\mathcal{B}_2$ is defined by 
 \begin{equation}
 k_{12}\triangleq \displaystyle
\sum_{\substack{X_1,X_2\in 2^\Theta\\ X_1\cap X_2=\emptyset}} m_{1}(X_1) m_{2}(X_2)
 \end{equation}

\subsection{The Murphy's rule}

The Murphy's rule of combination \cite{Murphy_2000} is a commutative but not associative trade-off rule, denoted here with index $M$, drawn from \cite{Yager_1985,Dubois_1988}. It is a special case of convex combination of bbas $m_1(.)$ and $m_2(.)$ and consists actually in a simple arithmetic average of belief functions associated with $m_1(.)$ and $m_2(.)$. $\text{Bel}_M(.)$ is then given $\forall X\in 2^\Theta$ by:
\begin{equation*}
\text{Bel}_M(X)=\frac{1}{2}[\text{Bel}_1(X) + \text{Bel}_2(X)]
\end{equation*}

\subsection{The Smets' rule}

The Smets' rule of combination \cite{Smets_1994,Smets_2000} is the non-normalized version of the conjunctive consensus (equivalent to the  non-normalized version of Dempster's rule).  It is commutative and associative and allows positive mass on the null/empty set $\emptyset$ (i.e. open-world assumption). Smets' rule of combination of two independent (equally reliable) sources of evidence (denoted here by index $S$) is then trivially given by:

$$m_S(\emptyset)\equiv k_{12}=\displaystyle\sum_{\substack{X_1,X_2\in 2^\Theta\\ X_1\cap X_2=\emptyset}} m_{1}(X_1) m_{2}(X_2)$$

\noindent
and $ \forall (X\neq\emptyset) \in 2^\Theta$, by

$$m_S(X)=\displaystyle\sum_{\substack{X_1,X_2\in 2^\Theta \\ X_1\cap X_2=X}}m_{1}(X_1)m_{2}(X_2)$$
\subsection{The Yager's rule }

The Yager's rule of combination \cite{Yager_1983, Yager_1985,Yager_1987} admits that in case of conflict the result is not reliable, so that $k_{12}$ plays the role of an absolute discounting term added to the weight of ignorance. This commutative but not associative rule, denoted here by index $Y$ is given\footnote{$\Theta$ represents here the full ignorance $\theta_1\cup\theta_2\cup\ldots\cup\theta_n$ on the frame of discernment according the notation used in \cite{Shafer_1976}.} by $m_Y(\emptyset)=0$ and $ \forall X\in 2^\Theta, X\neq\emptyset,\! X\neq\Theta$ by
$$m_Y(X)=\displaystyle\sum_{\substack{X_1,X_2\in 2^\Theta\\ X_1\cap X_2=X}} m_1(X_1)m_2(X_2)$$
\noindent 
and when $X=\Theta$ by
$$m_Y(\Theta)=m_1(\Theta)m_2(\Theta) + \displaystyle\sum_{\substack{X_1,X_2\in 2^\Theta\\ X_1\cap X_2=\emptyset}} m_1(X_1)m_2(X_2)$$

\subsection{The Dubois \& Prade's rule}

The Dubois \& Prade's rule of combination \cite{Dubois_1988} admits that the two sources are reliable when they are not in conflict, but one of them is right when a conflict occurs. Then if one observes a value in set $X_1$ while the other observes this value in a set $X_2$, the truth lies in $X_1\cap X_2$ as long $X_1\cap X_2\neq \emptyset$. If $X_1\cap X_2=\emptyset$, then the truth lies in $X_1\cup X_2$ \cite{Dubois_1988}. According to this principle, the commutative (but not associative) Dubois \& Prade hybrid rule of combination, denoted here by index $DP$, which is a reasonable trade-off between precision and reliability, is defined
by $m_{DP}(\emptyset)=0$ and $ \forall X\in 2^\Theta, X\neq\emptyset$ by
\begin{equation}
m_{DP}(X)=\displaystyle\sum_{\substack{X_1,X_2\in 2^\Theta\\ X_1\cap X_2=X \\ X_1\cap X_2\neq\emptyset}} m_1(X_1)m_2(X_2)  + \displaystyle\sum_{\substack{X_1,X_2\in 2^\Theta\\ X_1\cup X_2=X \\X_1\cap X_2=\emptyset}} m_1(X_1)m_2(X_2)
\end{equation}

\subsection{The disjunctive rule}

The disjunctive rule of combination \cite{Dubois_1986b,Dubois_1988,Smets_1993a} is a  commutative and associative rule proposed by Dubois \& Prade in 1986 and denoted here by the index $\cup$. $m_{\cup}(.)$ is defined $\forall X\in 2^\Theta$ by $m_{\cup}(\emptyset)=0$ and $ \forall (X\neq\emptyset) \in 2^\Theta$ by
$$m_{\cup}(X)=\displaystyle\sum_{\substack{X_1,X_2\in 2^\Theta\\ X_1\cup X_2=X}} m_{1}(X_1) m_{2}(X_2)$$

The core of the belief function given by  $m_{\cup}$ equals the union of the cores of $\text{Bel}_1$ and $\text{Bel}_2$. This rule reflects the disjunctive consensus and is usually preferred when one knows that one of the sources $\mathcal{B}_1$ or $\mathcal{B}_2$ is mistaken but without knowing which one among $\mathcal{B}_1$ and $\mathcal{B}_2$. Because we assume equi-reliability of sources in this paper, this rule will not be discussed in the sequel.

\subsection{Unification of the rules (weighted operator)}

In the framework of Dempster-Shafer Theory (DST), an unified formula has been proposed recently by Lef\`evre, Colot and Vanoorenberghe in \cite{Lefevre_2002} to embed all the existing (and potentially forthcoming) combination rules (including the PCR1 combination rule presented in the next section) involving conjunctive consensus in the same general mechanism of construction. We recently discovered that actually such unification formula had been already proposed 10 years before by Inagaki \cite{Inagaki_1991} as reported in \cite{Sentz_2002}. This formulation is known as {\it{the Weighted Operator}} (WO) in literature \cite{Josang_2003}, but since these two approaches have been developed independently by Inagaki and Lef\`evre et al., it seems more judicious to denote it as ILCV formula instead to refer to its authors when necessary (ILCV beeing the acronym standing for Inagaki-Lef\`evre-Colot-Vannoorenberghe). The WO (ILCV unified fusion rule) is based on two steps.

\begin{itemize}
\item {\bf{Step 1}}: Computation of the total conflicting mass based on the conjunctive consensus
\begin{equation}
k_{12} \triangleq \displaystyle\sum_{\substack{X_1,X_2\in 2^\Theta\\ X_1\cap X_2=\emptyset}}m_1(X_1)m_2(X_2) \end{equation}
\item {\bf{Step 2}}: This step consists in the reallocation (convex combination) of the conflicting masses on $(X\neq\emptyset)\subseteq \Theta$ with some given coefficients $w_m(X)\in[0,1]$ such that $\sum_{X\subseteq \Theta} w_m(X)=1$ according to
$$m(\emptyset)= w_m(\emptyset)\cdot k_{12}$$
\noindent
and $\forall (X\neq\emptyset)\in 2^\Theta$
\begin{equation}
m(X) = [\displaystyle\sum_{\substack{X_1,X_2\in 2^\Theta\\ X_1\cap X_2=X}} m_1(X_1)m_2(X_2)] + w_m(X)k_{12}
\label{eq:ILCV}
\end{equation}
\end{itemize}

This WO can be easily generalized for the combination of $N\geq 2$ independent and equi-reliable sources of information as well for step 2 by substituting $k_{12}$ by 

$$k_{12\ldots N}\triangleq \displaystyle\sum_{\substack{X_1,\ldots,X_N\in 2^\Theta\\ X_1\cap\ldots\cap X_N=\emptyset}}\prod_{i=1,N}m_i(X_i)$$
\noindent
and for step 2 by deriving for all $(X\neq\emptyset)\in 2^\Theta$ the mass $m(X)$ by

\begin{equation*}
m(X) = [\displaystyle\sum_{\substack{X_1,\ldots,X_N\in 2^\Theta\\ X_1\cap\ldots\cap X_N=X}} \prod_{i=1,N}m_i(X_i)] + w_m(X)k_{12\ldots N}
\end{equation*}

The particular choice of the set of coefficients $w_m(.)$ provides a particular rule of combination.
Actually this nice and important general formulation shows there exists an infinite number of possible rules of combination. Some rules are then justified or criticized with respect to the other ones mainly on their ability to, or not to, preserve the associativity and commutativity properties of the combination. It can be easily shown in  \cite{Lefevre_2002}  that such general procedure provides all existing rules involving conjunctive consensus developed in the literature based on Shafer's model. We will show later how the PCR1 rule of combination can also be expressed as a special case of the WO.

\subsection{The weighted average operator (WAO)}

This operator has been recently proposed by Josang, Daniel and Vannoorenberghe in \cite{Josang_2003}. It is a particular case of WO where the weighting coefficients $w_m(A)$ are chosen as follows: $w_m(\emptyset)=0$ and $\forall A\in 2^\Theta\setminus\{\emptyset\}$,
$$w_m(A)=\frac{1}{N} \sum_{i=1}^{N} m_i(A)$$
\noindent
where $N$ is the number of independent sources to combine.

\subsection{The hybrid DSm rule}

The hybrid DSm rule of combination is a new powerful rule of combination emerged from the recent theory of plausible and paradoxist reasoning developed by Dezert and Smarandache, known as DSmT in literature. The foundations of DSmT are different from the DST foundations and DSmT covers potentially a wider class of  applications than DST especially for dealing with highly conflicting static or dynamic fusion problems. Due to space limitations, we will not go further into a detailed presentation of DSmT here. A deep presentation of DSmT can be found in \cite{DSmTBook_2004a}. The DSmT deals properly with the granularity of information and intrinsic vague/fuzzy nature of elements of the frame $\Theta$ to manipulate. The basic idea of DSmT is to define belief assignments on hyper-power set $D^\Theta$ (i.e. free Dedekind's lattice) and to integrate all integrity constraints (exclusivity and/or non-existential constraints) of the model, say $\mathcal{M}(\Theta)$, fitting with the problem into the rule of combination. This rule, known as hybrid DSm rule works for any model (including the Shafer's model) and for any level of conflicting information. Mathematically, the hybrid DSm rule of combination of $N$ independent sources of evidence is defined as follows (see chap. 4 in \cite{DSmTBook_2004a}) for all $X\in D^\Theta$
\begin{equation}
m_{\mathcal{M}(\Theta)}(X)\triangleq 
\phi(X)\Bigl[ S_1(X) + S_2(X) + S_3(X)\Bigr]
 \label{eq:DSmHkBis1}
\end{equation}
\noindent
where $\phi(X)$ is the {\it{characteristic non-emptiness function}} of a set $X$, i.e. $\phi(X)= 1$ if  $X\notin \boldsymbol{\emptyset}$ and $\phi(X)= 0$ otherwise, where $\boldsymbol{\emptyset}\triangleq\{\boldsymbol{\emptyset}_{\mathcal{M}},\emptyset\}$. $\boldsymbol{\emptyset}_{\mathcal{M}}$ is the set  of all elements of $D^\Theta$ which have been forced to be empty through the constraints of the model $\mathcal{M}$ and $\emptyset$ is the classical/universal empty set. $S_1(X)$, $S_2(X)$ and $S_3(X)$ are defined by 
\begin{equation}
S_1(X)\triangleq \sum_{\substack{X_1,X_2,\ldots,X_N\in D^\Theta\\ (X_1\cap X_2\cap\ldots\cap X_N)=X}} \prod_{i=1}^{N} m_i(X_i)
\end{equation}
\begin{equation}
S_2(X)\triangleq \sum_{\substack{X_1,X_2,\ldots,X_N\in\boldsymbol{\emptyset}\\ [\mathcal{U}=X]\vee [(\mathcal{U}\in\boldsymbol{\emptyset}) \wedge (X=I_t)]}} \prod_{i=1}^{N} m_i(X_i)
\end{equation}
\begin{equation}
S_3(X)\triangleq\sum_{\substack{X_1,X_2,\ldots,X_N\in D^\Theta \\ (X_1\cup X_2\cup\ldots\cup X_N)=X \\ (X_1\cap X_2\cap \ldots\cap X_N)\in\boldsymbol{\emptyset}}}  \prod_{i=1}^{N} m_i(X_i)
\end{equation}
with $\mathcal{U}\triangleq u(X_1)\cup u(X_2)\cup \ldots \cup u(X_N)$ where $u(X_i)$, $i=1,\ldots,N$, is the union of all singletons $\theta_k$, $k\in\{1,\ldots,\vert\Theta\vert\}$, that compose $X_i$ and $I_t \triangleq \theta_1\cup \theta_2\cup\ldots\cup \theta_n$ is the total ignorance. $S_1(X)$ corresponds to the conjunctive consensus on free Dedekind's lattice for $N$ independent sources; $S_2(X)$ represents the mass of all relatively and absolutely empty sets which is transferred to the total or relative ignorances; $S_3(X)$ transfers the sum of relatively empty sets to the non-empty sets.\\

In the case of a dynamic fusion problem, when all elements become empty because one gets new evidence on integrity constraints (which corresponds to a specific hybrid model $\mathcal{M}$), then the conflicting mass is transferred to the total ignorance, which also turns to be empty, therefore 
the empty set gets now mass which means open-world, i.e, new hypotheses might 
be in the frame of discernment. For example, Let's consider the frame $\Theta=\{A,B\}$ with the 2 following bbas $m_1(A)=0.5$, $m_1(B)=0.3$, $m_1(A\cup B)=0.2$ and $m_2(A)=0.4$, $m_2(B)=0.5$, $m_2(A\cup B)=0.1$, but one finds out with new evidence that $A$ and $B$ are truly empty, then 
$A\cup B\equiv\Theta\overset{\mathcal{M}}{\equiv}\emptyset$. Then $m(\emptyset)=1$.\\

The hybrid DSm rule of combination is not equivalent to Dempter's rule even working on the Shafer's model. DSmT is actually a natural extension of the DST. An extension of this rule for the combination of {\it{imprecise}} generalized (or eventually classical) basic belief functions is possible and is presented in \cite{DSmTBook_2004a}.

\section{The PCR1 combination rule}

\subsection{The PCR1 rule for 2 sources}
\label{SEC:PCR1-2 sources}
Let $\Theta=\{\theta_1,\theta_2\}$ be the frame of discernment and its hyper-power set $D^\Theta = \{\emptyset,\theta_1, \theta_2, \theta_1\cup\theta_2\, \theta_1\cap\theta_2\}$.  Two basic belief assignments / masses $m_1(.)$ and $m_2(.)$ are defined over this hyper-power set.   
We assume that $m_1(.)$ and $m_2(.)$ are normalized belief masses following definition given by \eqref{eq:bba}. The PCR1 combination rule consists in two steps:

\begin{itemize}
\item {\bf{Step 1}}: Computation of the conjunctive consensus\footnote{$\oplus$ denotes here the generic symbol for the fusion.} $m_\cap(.)=[m_1\oplus m_2](.)$ and the conflicting mass according to
\begin{equation}
m_\cap(X)=\sum_{\substack{X_1,X_2\in D^\Theta\\ X_1\cap X_2=X}} m_1(X_1)m_2(X_2)
 \end{equation}
\noindent
and
\begin{equation}
k_{12} \triangleq \displaystyle\sum_{\substack{X_1,X_2\in D^\Theta\\ X_1\cap X_2=\emptyset}}m_1(X_1)m_2(X_2) \end{equation}

This step coincides with the Smets' rule of combination when accepting the open-world assumption.
In the Smets' open-world TBM framework \cite{Smets_1990}, $k_{12}$ is interpreted as the mass $m(\emptyset)$ committed to the empty set. $\emptyset$ corresponds then to all missing unknown hypotheses and the absolute impossible event. 

\item {\bf{Step 2}} (normalization): Distribution of  the conflicting mass $k_{12}$ onto $m_\cap(X)$ proportionally with the non-zero sums of their corresponding columns of non-empty sets of the effective mass matrix $\mathbf{M}_{12}[m_{ij}]$ (index $12$ denotes the list of sources entering into the mass matrix).  If all sets are empty, then the conflicting mass is redistributed to the disjunctive form of all these empty sets (which is many cases coincides with the total ignorance).\\

More precisely, the original mass matrix $\mathbf{M}_{12}$ is a $(N=2)\times (2^{\vert \Theta \vert}-1)$ matrix constructed by stacking the row vectors 
\begin{equation*}
\begin{cases}
\mathbf{m}_1=[m_1(\theta_1) \quad m_1(\theta_2)  \quad m_1(\theta_1\cup \theta_2)]\\
\mathbf{m}_2=[m_2(\theta_1)  \quad m_2(\theta_2)  \quad m_2(\theta_1\cup \theta_2)]
\end{cases}
\end{equation*}
associated with the beliefs assignments $m_1(.)$ and $m_2(.)$. For convenience and by convention, the row index $i$ follows the index of sources and the index $j$ for columns follows the enumeration of elements of power set $2^\Theta$ (excluding the empty set because by definition its committed mass is zero). Any permutation of rows and columns can be arbitrarily chosen as well and it doesn't not make any difference in the PCR1 fusion result. Thus, one has for the 2 sources and 2D fusion problem:
\begin{equation*}
\mathbf{M}_{12}=
\begin{bmatrix}
\mathbf{m}_1\\
\mathbf{m}_2
\end{bmatrix}
=
\begin{bmatrix}
m_1(\theta_1) & m_1(\theta_2) & m_1(\theta_1\cup \theta_2)\\
m_2(\theta_1) & m_2(\theta_2) & m_2(\theta_1\cup \theta_2)
\end{bmatrix}
\end{equation*}

We denote by $c_{12}(X)$ the sum of the elements of the column of the mass matrix associated with element $X$ of the power set, i.e
\begin{equation*}
\begin{cases}
c_{12}(X=\theta_1)=m_1(\theta_1)+m_2(\theta_1)\\
c_{12}(X=\theta_2)=m_1(\theta_2)+m_2(\theta_2)\\
c_{12}(X=\theta_1\cup\theta_2)=m_1(\theta_1\cup\theta_2)+m_2(\theta_1\cup\theta_2)
\end{cases}
\end{equation*}

The conflicting mass $k_{12}$ is distributed proportionally with all non-zero coefficients $c_{12}(X)$. For elements $X\in D^\Theta$ with zero coefficients $c_{12}(X)$, no conflicting mass will be distributed to them. Let's note by $w(\theta_1)$, $w(\theta_2)$ and $w(\theta_1\cup\theta_2)$ the part of the conflicting mass that is respectively distributed to $\theta_1$, $\theta_2$ and $\theta_1\cup\theta_2$ (assuming $c_{12}(\theta_1)>0$, $c_{12}(\theta_2)>0$ and $c_{12}(\theta_1\cup\theta_2)>0$. Then:
\begin{equation}
\frac{w(\theta_1)}{c_{12}(\theta_1)}=\frac{w(\theta_2)}{c_{12}(\theta_2)}=\frac{w(\theta_1\cup\theta_2)}{c_{12}(\theta_1\cup\theta_2)}
=\frac{w(\theta_1)+w(\theta_2)+w(\theta_1\cup\theta_2)}{c_{12}(\theta_1)+c_{12}(\theta_2)+c_{12}(\theta_1\cup\theta_2)}=\frac{k_{12}}{d_{12}}
\end{equation}
\noindent 
because 
\begin{equation*}
c_{12}(\theta_1)+c_{12}(\theta_2)+c_{12}(\theta_1\cup\theta_2)\\
=\sum_{X_1\in D^\Theta \setminus\{\emptyset\}}m_1(X_1)+\sum_{X_2\in D^\Theta \setminus\{\emptyset\}}m_2(X_2)=d_{12}
\end{equation*}
\noindent
Hence the proportionalized conflicting masses to transfer are given by

\begin{equation*}
\begin{cases}
w(\theta_1)=c_{12}(\theta_1)\cdot \frac{k_{12}}{d_{12}}\\
w(\theta_2)=c_{12}(\theta_2)\cdot \frac{k_{12}}{d_{12}}\\
w(\theta_1\cup\theta_2)=c_{12}(\theta_1\cup\theta_2)\cdot \frac{k_{12}}{d_{12}}
\end{cases}
\end{equation*}

\noindent 
which are added respectively to $m_\cap(\theta_1)$, $m_\cap(\theta_2)$ and $m_\cap(\theta_1\cup\theta_2)$.

\end{itemize}

Therefore, the general formula for the PCR1 rule for 2 sources, for $\vert \Theta\vert \geq 2$, is given by $m_{PCR1}(\emptyset)=0$ and for $(X\neq\emptyset)\in D^\Theta$,
\begin{equation}
m_{PCR1}(X)=\sum_{\substack{X_1,X_2\in D^\Theta\\ X_1\cap X_2=X}} m_1(X_1)m_2(X_2) + c_{12}(X)\cdot \frac{k_{12}}{d_{12}}
\label{eq:PCR22D}
\end{equation}

\noindent
where $k_{12}$ is the total conflicting mass and $c_{12}(X)\triangleq \sum_{i=1,2} m_i(X) \neq 0$, i.e. the non-zero sum of the column of the mass matrix $\mathbf{M}_{12}$ corresponding to the element $X$, and $d_{12}$ is the sum of all non-zero column sums of all non-empty sets (in many cases $d_{12}=2$ but in some degenerate cases it can be less).\\

In the degenerate case when all column sums of all non-empty sets are zero, then the conflicting mass is transferred to the non-empty disjunctive form of all sets involved in the conflict together.  But if this disjunctive form happens to be empty, then one considers an open world (i.e. the frame of discernment might contain new hypotheses) and thus all conflicting mass is transferred to the empty set.\\

As seen, the PCR1 combination rule works for any degree of conflict $k_{12}\in [0,1]$, while Dempster's rule does not work for $k_{12}=1$ and gives counter-intuitive results for most of high conflicting fusion problems.

\subsection{Generalization for $N\geq 2$ sources}

The previous PCR1 rule of combination for two sources ($N=2$) can be directly and easily extended for the multi-source case ($N\geq 2$) as well. The general formula of the PCR1 rule is thus given by $m_{PCR1}(\emptyset)=0$ and for $X\neq\emptyset)\in D^\Theta$
\begin{equation}
m_{PCR1}(X)=\bigl[\sum_{\substack{X_1,\ldots,X_N\in D^\Theta\\ X_1\cap\ldots\cap X_N=X}} \prod_{i=1,N}m_i(X_i)\bigr]\\
 + c_{12\ldots N}(X)\cdot \frac{k_{12\ldots N}}{d_{12\ldots N}}
\label{eq:PCRn}
\end{equation}

\noindent
where $k_{12\ldots N}$ is the total conflicting mass between all the $N$ sources which is given by
\begin{equation}
k_{12\ldots N} \triangleq \displaystyle\sum_{\substack{X_1,\ldots ,X_N\in D^\Theta\\ X_1\cap\ldots\cap X_N=\emptyset}}\prod_{i=1,N}m_i(X_i) 
\end{equation}
\noindent
and $c_{12\ldots N}(X)\triangleq \sum_{i=1,N} m_i(X) \neq 0$, i.e. the non-zero sum of the column of the mass matrix $\mathbf{M}_{12\ldots N}$ corresponding to the element $X$, while $ d_{12\ldots N}$ represents the sum of all non-zero column sums of all non-empty sets (in many cases $d_{12\ldots N}=N$ but in some degenerate cases it can be less).\\

Similarly for $N$ sources, in the degenerate case when all column sums of all non-empty sets are zero, then the conflicting mass is transferred to the non-empty disjunctive form of all sets involved in the conflict together.  But if this disjunctive form happens to be empty, then one considers an open world (i.e. the frame of discernment might contain new hypotheses) and thus all conflicting mass is transferred to the empty set.\\

The PCR1 rule can be seen as a cheapest, easiest implementable approximated version of the sophisticated MinC combination rule proposed by Daniel in \cite{Daniel_2003} and \cite{DSmTBook_2004a} (chap. 10). Note also that the PCR1 rule works in the DSmT framework and can serve as a cheap alternative to the more sophisticated and specific DSm hybrid rule but preferentially when none of sources is totally ignorant (see discussion in section 3.6).  One applies the DSm classic rule \cite{DSmTBook_2004a} (i.e. the conjunctive consensus on $D^\Theta$), afterwards one identifies the model and its integrity constraints and one eventually employs the PCR1 rule instead of DSm hybrid rule (depending of the dimension of the problem to solve, the number of sources involved and the computing resources available). PCR1 can be used on the power set $2^\Theta$ and within the DS Theory. \\

The PCR1 combination rule is commutative but not associative. It converges towards Murphy's rule (arithmetic mean of masses) when the conflict is approaching 1, and it converges towards the conjunctive consensus rule when the conflict is approaching 0.

\subsection{Implementation of the PCR1 rule}

For practical use and implementation of the PCR1 combination rule, it is important to save memory space and avoid useless computation as best as possible and especially when dealing with many sources and for frames of high dimension. To achieve this, it's important to note that since all zero-columns of the mass matrix do not play a role in the normalization, all zero-columns (if any) of the original mass matrix can be removed to {\it{compress}} the matrix horizontally (this can be easily done using MatLab programming language) to get an effective mass matrix of smaller dimension for computation the set of proportionalized conflicting masses to transfer. The list of elements of power set corresponding to non-empty colums must be maintained in parallel to this compression for implementation purpose. By example, let's assume $\vert\Theta\vert = 2$ and only 2 sources providing  $m_1(\theta_2)=m_2(\theta_2)=0$ and all other masses are positive, then the {\it{effective}} mass matrix will become

\begin{equation*}
\mathbf{M}_{12}=
\begin{bmatrix}
m_1(\theta_1) & m_1(\theta_1\cup \theta_2)\\
m_2(\theta_1) & m_2(\theta_1\cup \theta_2)
\end{bmatrix}
\end{equation*}
\noindent
with now the following correspondance for column indexes: $(j=1)\leftrightarrow \theta_1$ and $(j=2)\leftrightarrow \theta_1\cup\theta_2$.\\

The computation the set of proportionalized conflicting masses to transfer will be done using the PCR1 general formula directly from this previous effective mass matrix rather than from
\begin{equation*}
\mathbf{M}_{12}=
\begin{bmatrix}
\mathbf{m}_1\\
\mathbf{m}_2
\end{bmatrix}
=
\begin{bmatrix}
m_1(\theta_1) & m_1(\theta_2)=0 & m_1(\theta_1\cup \theta_2)\\
m_2(\theta_1) & m_2(\theta_2)=0 & m_2(\theta_1\cup \theta_2)
\end{bmatrix}
\end{equation*}

\subsection{PCR1 rule as a special case of WO}

The PCR1 rule can be easily expressed as a special case of the WO \eqref{eq:ILCV} for the combination of two sources by choosing as weighting coefficients for each $X\in 2^\Theta \setminus\{ \emptyset\}$,
$$w_m(X)=c_{12}(X)/{d_{12}}$$
For the combination of $N\geq 2$ independent and equi-reliable sources, the weighting coefficients will be given by
$$w_m(X)=c_{12\ldots N}(X)/{d_{12 \dots N}}$$

\subsection{Advantages of the PCR1 rule}

\begin{itemize}
\item the PCR1 rule works in any cases, no matter what the conflict is (it may be 1 or less); Zadeh's example, examples with $k_{12}=1$ or $k_{12}=0.99$, etc. All work;
\item the implementation of PCR1 rule is very easy and thus presents a great interest for engineers who look for a cheap and an easy alternative fusion rule to existing rules;
\item  the PCR1 formula is simple (it is not necessary to go by proportionalization each time when fusionning);
\item the PCR1 rule works quite well with respect to some other rules since the specificity of information is preserved (i.e no mass is transferred onto partial or total ignorances, neither onto the empty set as in TBM);
\item the PCR1 rule reflects the majority rule;
\item the PCR1 rule is convergent towards idempotence for problems with no unions or intersections of sets (we know that, in fact, no combination rule is idempotent, except Murphy elementary fusion mean rule);
\item the PCR1 rule is similar to the classical Dempster-Shafer's rule instead of proportionalizing with respect to the results of the conjunctive rule as is done in Dempster's, we proportionalize with respect to the non-zero sum of the columns masses, the only difference is that in the DS combination rule one eliminates the denominator (which caused problems when the degree of conflict is 1 or close to 1); PCR1 on the power set and for non-degenerate cases gives the same results as WAO \cite{Josang_2003}; yet, for the storage proposal in a dynamic fusion when the associativity is needed, for PCR1 is needed to store only the last sum of masses, besides the previous conjunctive rules result, while in WAO it is in addition needed to store the number of the steps and both rules become quasi-associative; 
\item the normalization, done proportionally with the corresponding non-zero sum of elements of the mass matrix, is natural - because the more mass is assigned to an hypothesis by the sources the more mass that hypothesis deserves to get after the fusion.
\end{itemize}

\subsection{Disadvantages of the PCR1 rule}

\begin{itemize}
\item the PCR1 rule requires normalization/proportionalization, but the majority of rules do; rules which do not require normalization loose information through the transfer of conflicting mass to partial and/or total ignorances or to the empty set.
\item the results of PCR1 combination rule do not bring into consideration any new set: formed by unions (uncertainties); or intersections (consensus between some hypotheses); yet, in the DSmT framework the intersections show up through the hyper-power set.
\item the severe drawback of PCR1 and WAO rules is that they do not preserve the neutrality property of the vacuous belief assignment $m_v(.)$ (defined by $m_v(\Theta)=1$) as one legitimately expects since 
if one or more bbas $m_s(.)$, $s\geq 1$, different from
the vacuous belief, are combined with the vacuous belief assignment the result is not the same as that
of the combination of the bbas only (without including $m_v(.)$), i.e. $m_v(.)$ does not act as a neutral element for the fusion combination.
%

%
In other words, for $s\geq 1$, one gets for $m_1(.)\neq m_v(.)$, \ldots, $m_s(.)\neq m_v(.)$:
\begin{equation}
m_{PCR1}(.)=[m_1\oplus\ldots m_s\oplus m_v](.) \neq [m_1\oplus\ldots m_s](.)
\end{equation}
\begin{equation}
m_{WAO}(.)=[m_1\oplus\ldots m_s\oplus m_v](.)\neq [m_1\oplus\ldots m_s](.)
\end{equation}
For the cases of the combination of only one non-vacuous belief assignment $m_1(.)$ with the vacuous belief assignment $m_v(.)$ where $m_1(.)$ has mass asigned to an empty element, say
$m_1(\emptyset)>0$ as in Smets' TBM, or as in DSmT dynamic
fusion where one finds out that a previous non-empty element
$A$, whose mass $m_1(A)>0$, becomes empty after a certain
time, then this mass of an empty set has to be
transferred to other elements using PCR1, but for such case $[m_1\oplus m_v](.)]$ is different from $m_1(.)$.

\noindent
{\bf{Example:}}  Let's have $\Theta=\{A,B\}$ and two bbas 
$$m_1(A)=0.4\qquad m_1(B)=0.5\qquad m_1(A\cup B)=0.1$$
$$m_2(A)=0.6\qquad m_2(B)=0.2\qquad m_2(A\cup B)=0.2$$
\noindent
together with the vacuous bba $m_v(\Theta=A\cup B)=1$.
If one applies the PCR1 rule to combine the 3 sources altogether, one gets
$$m_{PCR1|12v}(A)= 0.38 + 1\cdot \frac{0.38}{3} = 0.506667$$
$$m_{PCR1|12v}(B)= 0.22 + 0.7\cdot \frac{0.38}{3}  = 0.308667$$
$$m_{PCR1|12v}(A\cup B)=  0.02 + 1.3\cdot \frac{0.38}{3}  = 0.184666$$
\noindent 
since the conjunctive consensus is given by
$m_{12v}(A)=0.38$, $m_{12v}(B)=0.22$, $m_{12v}(A\cup B)=0.02$; the conflicting mass is $k_{12v}=0.38$ and one has
$$\frac{x}{1}=\frac{y}{0.7}=\frac{z}{1.3}=\frac{0.38}{3}$$
\noindent while the combination of only the sources 1 and 2 withe the PCR1 provides
$$m_{PCR1|12}(A)= 0.38 + 0.19 = 0.570$$
$$m_{PCR1|12}(B)= 0.22 + 0.133 = 0.353$$
$$m_{PCR1|12}(A\cup B)=  0.02+0.057= 0.077$$
since the conjunctive consensus is given by
$m_{12}(A)=0.38$, $m_{12}(B)=0.22$, $m_{12}(A\cup B)=0.02$; the conflicting mass is $k_{12}=0.38$ but one has now the following redistribution condition
$$\frac{x}{1}=\frac{y}{0.7}=\frac{z}{0.3}=\frac{0.38}{2}=0.19$$
Thus clearly $m_{PCR1|12v}(.)\neq m_{PCR1|12}(.)$ although the third source brings no information in the fusion since it is fully ignorant. This behavior is abnormal and counter-rintuitive. WAO gives the same results in this example, therefore WAO also doesn't satisfy the neutrality property of the vacuous belief assignment for the fusion. That's why we have improved PCR1 to PCR2-4 rules in a companion paper \cite{DSmTBook_2004b}.

%

\end{itemize}

\subsection{Comparison of the PCR1 rule with the WAO}

\subsubsection{The non degenerate case}

Let's compare in this section the PCR1 with the WAO for a very simple 2D general non degenerate case (none of the elements of the power set or hyper-power set of the frame $\Theta$ are known to be truly empty but the universal empty set itself) for the combination of 2 sources. Assume that the non degenerate mass matrix $\mathbf{M}_{12}$ associated with the beliefs assignments $m_1(.)$ and $m_2(.)$ is given by 
\begin{equation*}
\begin{cases}
\mathbf{m}_1=[m_1(\theta_1) \quad m_1(\theta_2)  \quad m_1(\theta_1\cup \theta_2)]\\
\mathbf{m}_2=[m_2(\theta_1)  \quad m_2(\theta_2)  \quad m_2(\theta_1\cup \theta_2)]
\end{cases}
\end{equation*}

In this very simple case, the total conflict is given by

$$k_{12}=m_1(\theta_1)m_2(\theta_2)+m_1(\theta_2)m_2(\theta_1)$$

According to the WAO definition, one gets $m_{WAO}(\emptyset)=w_m(\emptyset)\cdot k_{12} = 0$ because by definition $w_m(\emptyset)=0$. The other weighting coefficients of WAO are given by

$$w_m(\theta_1)=\frac{1}{2}[m_1(\theta_1)+m_2(\theta_1)]$$
$$w_m(\theta_2)=\frac{1}{2}[m_1(\theta_2)+m_2(\theta_2)]$$
$$w_m(\theta_1\cup \theta_2)=\frac{1}{2}[m_1(\theta_1\cup \theta_2)+m_2(\theta_1\cup \theta_2)]$$

\noindent
Thus, one obtains
\begin{multline*}
m_{WAO}(\theta_1)=[m_1(\theta_1)m_2(\theta_1)+m_1(\theta_1\cup \theta_2)m_2(\theta_1)+ m_1(\theta_1)m_2(\theta_1\cup \theta_2)] \\ + \frac{1}{2}[m_1(\theta_1)+m_2(\theta_1)]
\cdot [m_1(\theta_1)m_2(\theta_2)+m_1(\theta_2)m_2(\theta_1)]
\end{multline*}
\begin{multline*}
m_{WAO}(\theta_2)=[m_1(\theta_2)m_2(\theta_2)+m_1(\theta_1\cup \theta_2)m_2(\theta_2)+ m_1(\theta_2)m_2(\theta_1\cup \theta_2)] \\
+ \frac{1}{2}[m_1(\theta_2)+m_2(\theta_2)]
\cdot [m_1(\theta_1)m_2(\theta_2)+m_1(\theta_2)m_2(\theta_1)]
\end{multline*}
\begin{equation*}
m_{WAO}(\theta_1\cup \theta_2)=[m_1(\theta_1\cup \theta_2)m_2(\theta_1\cup \theta_2)]
+ \frac{1}{2}[m_1(\theta_1\cup \theta_2)+m_2(\theta_1\cup \theta_2)]
\cdot [m_1(\theta_1)m_2(\theta_2)+m_1(\theta_2)m_2(\theta_1)]
\end{equation*}

\noindent
It is easy to verify that  $\sum_{X\in 2^\Theta}m_{WAO}(X)=1$.\\

Using the PCR1 formula for 2 sources explicated in section \ref{SEC:PCR1-2 sources}, one has $m_{PCR1}(\emptyset)=0$ and the weighting coefficients of the PCR1 rule are given by
\begin{equation*}
\begin{cases}
c_{12}(\theta_1)=m_1(\theta_1)+m_2(\theta_1)\\
c_{12}(\theta_2)=m_1(\theta_2)+m_2(\theta_2)\\
c_{12}(\theta_1\cup\theta_2)=m_1(\theta_1\cup\theta_2)+m_2(\theta_1\cup\theta_2)
\end{cases}
\end{equation*}
and $d_{12}$ by $d_{12}=c_{12}(\theta_1)+c_{12}(\theta_2)+c_{12}(\theta_1\cup\theta_2)=2$.
Therefore, one finally gets:
\begin{multline*}
m_{PCR1}(\theta_1)=[m_1(\theta_1)m_2(\theta_1)+m_1(\theta_1\cup \theta_2)m_2(\theta_1)+ m_1(\theta_1)m_2(\theta_1\cup \theta_2)] \\
+ \frac{c_{12}(\theta_1)}{d_{12}}
\cdot [m_1(\theta_1)m_2(\theta_2)+m_1(\theta_2)m_2(\theta_1)]
\end{multline*}
\begin{multline*}
m_{PCR1}(\theta_2)=[m_1(\theta_2)m_2(\theta_2)+m_1(\theta_1\cup \theta_2)m_2(\theta_2)+ m_1(\theta_2)m_2(\theta_1\cup \theta_2)] \\
+  \frac{c_{12}(\theta_2)}{d_{12}}
\cdot [m_1(\theta_1)m_2(\theta_2)+m_1(\theta_2)m_2(\theta_1)]
\end{multline*}
\begin{equation*}
m_{PCR1}(\theta_1\cup \theta_2)=[m_1(\theta_1\cup \theta_2)m_2(\theta_1\cup \theta_2)] 
+  \frac{c_{12}(\theta_1\cup\theta_2)}{d_{12}}
\cdot [m_1(\theta_1)m_2(\theta_2)+m_1(\theta_2)m_2(\theta_1)]
\end{equation*}

Therefore for all $X$ in $2^\Theta$, one has $m_{PCR1}(X)=m_{WAO}(X)$ if no singletons or unions of singletons are (or become) empty at a given time, otherwise the results are different as seen in the below three examples. This property holds for the combination of $N > 2$ sources working on a $n-D$ frame ($n>2$) $\Theta$ as well if no singletons or unions of singletons are (or become) empty at a given time, otherwise the results become different.

\subsubsection{The degenerate case}

In the dynamic fusion, when one or more singletons or unions of 
singletons become empty at a certain time $t$ which corresponds to a degenerate case, the WAO does not work.\\

\noindent
{\bf{Example 1}}:  Let's consider the Shafer's model (exhaustivity and exclusivity of hypotheses) on $\Theta=\{A,B,C\}$ and the two following bbas 
$$m_1(A)=0.3 \quad m_1(B)=0.4\quad m_1(C)=0.3$$
$$m_2(A)=0.5\quad m_2(B)=0.1\quad m_2(C)=0.4$$
\noindent 
Then the conjunctive consensus yields
$$m_{12}(A)=0.15\quad m_{12}(B)=0.04\quad m_{12}(C)=0.12$$
\noindent
and the conflicting mass $k_{12}=0.69$.
Now assume that at time $t$, one finds out that $B=\emptyset$, then the new conflict mass which becomes $k_{12}'=0.69+0.04=0.73$ is re-distributed to $A$ and $B$ according to the WAO formula:
$$m_{WAO}(B) = 0$$
$$m_{WAO}(A) = 0.15 + (1/2)(0.3+0.5)(0.73) = 0.4420$$
$$m_{WAO}(C) = 0.12 + (1/2)(0.3+0.4)(0.73) = 0.3755$$

\noindent
From this WAO result, one sees clearly that the sum of the combined masses $m(.)$ is $0.8175 < 1$ while using PCR1, one redistributes $0.73$ to $A$ and $B$ following the PCR1 formula:
$$m_{PCR1}(B) = 0$$
\begin{equation*}
m_{PCR1}(A) = 0.15 + \frac{(0.3+0.5)(0.73)}{(0.3+0.5 + 0.3+0.4)} = 0.539333
\end{equation*}
\begin{equation*}
m_{PCR1}(C) = 0.12 + \frac{(0.3+0.4)(0.73)}{(0.3+0.5 + 0.3+0.4)}= 0.460667
\end{equation*}
\noindent
which clearly shows that he sum of masses $m_{PCR1}(.)$ is 1 as expected for a proper belief assignment.\\

\noindent
{\bf{Example 2}} (totally degenerate case) :  Let's take exactly the same previous example with exclusive hypotheses $A$, $B$ and $C$ but assume now that at time $t$ one finds out that $A$, $B$ and $C$ are all truly empty, then $k_{12}'=1$. In this case, the WAO is not able to redistribute the conflict to any element $A$, $B$, $C$  or partial/total ignorances because they are empty. But PCR1 transfers the conflicting mass to the ignorance $A\cup B \cup C$, which is the total ignorance herein, but this is also empty, thus the conflicting mass is transferred to the empty set, meaning we have an open world, i.e. new hypotheses might belong to the frame of discernment.\\

\noindent

{\bf{Example 3}} (Open-world):  In the Smets' open-world approach (when the empty set gets some mass assigned by the sources), the WAO doesn't work either. For example, let's consider $\Theta=\{A,B\}$ and the following bbas $m_1(\emptyset)=0.1$, $m_2(\emptyset)=0.2$ and
$$m_1(A)=0.4\quad m_1(B)=0.3\quad m_1(A\cup B)=0.2$$
$$m_2(A)=0.5\quad m_2(B)=0.2\quad m_2(A\cup B)=0.1$$
Then the conjunctive consensus yields $m_{12}(\emptyset)=0.28$ and $$m_{12}(A)=0.34\quad m_{12}(B)=0.13 \quad m_{12}(A\cup B)=0.02$$ with the conflicting mass 
$$k_{12}=m_{12}(A\cap B) + m_{12}(\emptyset)=0.23+0.28=0.51$$

\noindent
Using WAO, one gets
$$m_{WAO}(\emptyset)=0$$
$$m_{WAO}(A)= 0.34 + (1/2)(0.4+0.5)(0.51) = 0.5695$$
$$m_{WAO}(B)= 0.13 + (1/2)(0.3+0.2)(0.51) = 0.2275$$
$$m_{WAO}(A\cup B)= 0.02 + (1/2)(0.2+0.1)(0.51) =  0.0965$$
The sum of masses$m_{WAO}(.)$ is $0.9235 < 1$ while PCR1 gives:
$$m_{PCR1}(\emptyset)=0$$
\begin{equation*}
m_{PCR1}(A)= 0.34 + \frac{(0.4+0.5)(0.51)}{(0.4+0.5+0.3+0.2+0.2+0.1)}= 0.61
\end{equation*}
\begin{equation*}
m_{PCR1}(B) = 0.13 + \frac{(0.3+0.2)(0.51)}{(0.4+0.5+0.3+0.2+0.2+0.1)} = 0.28
\end{equation*}
\begin{equation*}
m_{PCR1}(A\cup B) = 0.02 +  \frac{(0.2+0.1)(0.51)}{(0.4+0.5+0.3+0.2+0.2+0.1)}= 0.11
\end{equation*}
\noindent
which shows that the sum of masses $m_{PCR1}(.)$ is 1.

\subsubsection{Comparison of memory storages}

In order to keep the associativity of PCR1 one stores the previous
result of combination using the conjunctive rule, and also the sums of mass
columns [2 storages]. For the WAO one stores the previous result of combination using the
conjunctive rule (as in PCR1), and the mass columns averages (but the
second one is not enough in order to compute the next average and
that's why one still needs to store the number of masses combined so far) [3
storages].\\

For example, let's $\Theta = \{A, B, C\}$ and let's suppose first that only five bbas available, $m_1(.)$, $m_2(.)$, $m_3(.)$, $m_4(.)$, $m_5(.)$, have been combined with WAO, where for example $m_1(A)=0.4$, $m_2(A)=0.2$, $m_3(A)=0.3$, $m_4(A)=0.6$, $m_5(A)=0.0$.  Their average $m_{12345}(A)=0.3$ was then obtained and stored.  Let's assume now that a new bba $m_6(.)$, with $m_6(A)=0.4$ comes in as a new 
evidence.  Then, how to compute with WAO the new average $m_{123456}(A)=[m_{12345}\oplus m_6](A)$?  We need to know how many masses have been combined so far with WAO (while in PCR1 this is not necessary). Therefore $n=5$, the number of combined bbas so far, has to 
be stored too when using WAO in sequential/iterative fusion.  Whence, the new average is possible to be computed with WAO :

$$m_{123456}(A)=\frac{5\cdot 0.3 + 0.4}{5+1} = 0.316667$$
\noindent 
but contrariwise to WAO, we don't need an extra memory storage for keep in memory $n=5$ when using PCR1 to compute\footnote{The notation $m_{PCR1|12\ldots n}(.)$ denotes explicitly the fusion of $n$ bbas $m_1(.)$, $m_2(.)$, \ldots, $m_n(.)$; i.e. given the knowledge of the $n$ bbas combined altogether. } $m_{PCR1|123456}(A)$ from $m_{PCR1|12345}(A)$ and $m_6(A)$ which is more interesting since PCR1 reduces the memory storage requirement versus WAO. Indeed, using PCR1 we only store the sum of previous masses:
$c_{12345}(A)=0.4+0.2+0.3+0.6+0.0=1.5$, and when another
bba $m_6(.)$ with $m_6(A)=0.4$ comes in as a new evidence one
only adds it to the previous sum of masses:
$c_{123456}(A)=1.5+0.4=1.9$ to get the coefficient of
proportionalization for the set $A$.

\section{Some numerical examples}

\subsection{Example 1}

Let's consider a general 2D case (i.e. $\Theta=\{\theta_1,\theta_2\}$) including epistemic uncertainties with the two following belief assignments
$$m_1(\theta_1)=0.6, \quad m_1(\theta_2)=0.3,\quad m_1(\theta_1\cup\theta_2)=0.1$$
$$m_2(\theta_1)=0.5,\quad m_2(\theta_2)=0.2,\quad m_2(\theta_1\cup\theta_2)=0.3$$
The conjunctive consensus yields:
$$m_\cap(\theta_1)=0.53,\quad m_\cap(\theta_2)=0.17,\quad m_\cap(\theta_1\cup\theta_2)=0.03$$
\noindent 
with the total conflicting mass $k_{12}=0.27$.\\

Applying the proportionalization from the mass matrix
\begin{equation*}
\mathbf{M}_{12}=
\begin{bmatrix}
0.6 & 0.3 & 0.1\\
0.5 & 0.2 & 0.3
\end{bmatrix}
\end{equation*}
\noindent one has
\begin{equation*}
\frac{w_{12}(\theta_1)}{0.6+0.5}=\frac{w_{12}(\theta_2)}{0.3+0.2}=\frac{w_{12}(\theta_1\cup\theta_2)}{0.1+0.3}=\frac{w_{12}(\theta_1)+w_{12}(\theta_2)+w_{12}(\theta_1\cup\theta_2)}{2}=\frac{0.27}{2}=0.135
\end{equation*}
\noindent
and thus one deduces:
$$w_{12}(\theta_1)=1.1\cdot 0.135=0.1485\qquad w_{12}(\theta_2) = 0.5\cdot 0.135=0.0675\qquad w_{12}(\theta_1\cup\theta_2) = 0.4\cdot 0.135=0.0540$$ 
One adds $w_{12}(\theta_1)$ to $m_\cap(\theta_1)$, $w_{12}(\theta_2)$ to $m_\cap(\theta_2)$ and $w_{12}(\theta_1\cup\theta_2)$ to $m_\cap(\theta_1\cup\theta_2)$. One finally gets the result of the PCR1 rule of combination:
$$m_{PCR1}(\theta_1)= 0.53+0.1485=0.6785$$ $$m_{PCR1}(\theta_2)=0.17+0.0675=0.2375$$  $$m_{PCR1}(\theta_1\cup\theta_2)=0.03+0.0540=0.0840$$

\subsection{Example 2}
Let's consider the frame of discernment with only two exclusive elements, i.e. $\Theta=\{\theta_1,\theta_2\}$ and consider the two following {\it{Bayesian}} belief assignments
$$m_1(\theta_1)=0.2, m_1(\theta_2)=0.8$$
$$m_2(\theta_1)=0.9, m_2(\theta_2)=0.1$$

The associated (effective) mass matrix will be

\begin{equation*}
\mathbf{M}_{12}=
\begin{bmatrix}
0.2 & 0.8\\
0.9 & 0.1
\end{bmatrix}
\end{equation*}

The first row of $\mathbf{M}_{12}$ corresponds to basic belief assignment $m_1(.)$ and the 
second row of $\mathbf{M}_{12}$ corresponds to basic belief assignment $m_2(.)$. The columns of the mass matrix $\mathbf{M}_{12}$ correspond to focal elements of $m_1(.)$ and $m_2(.)$ and the choice for ordering these elements doesn't matter. any arbitrary choice is possible. In this example the first column of $\mathbf{M}_{12}$ is associated with $\theta_1$ and the second column with $\theta_2$.

\subsubsection{Fusion with the PCR1 rule}

The conjunctive consensus yields:
$$\begin{cases}
m_\cap(\theta_1)=[m_1\oplus m_2](\theta_1)=0.2\cdot 0.9=0.18\\
m_\cap(\theta_2)=[m_1\oplus m_2](\theta_2)=0.8\cdot 0.1=0.08
\end{cases}$$

\noindent
The remaining mass corresponds to the conflict $k_{12}$, i.e. 
\begin{equation*}
k_{12} = 1-m_\cap(\theta_1)-m_\cap(\theta_2)= m_1(\theta_1)m_2(\theta_2)+m_1(\theta_2)m_2(\theta_1)=(0.2\cdot 0.1)+(0.9\cdot 0.8)=0.74
\end{equation*}

Now the conflicting mass, $k_{12}=0.74$, is distributed between $m_\cap(\theta_1)$ and $m_\cap(\theta_2)$ proportionally with the non-zero sums of their columns.  Thus, the column vector associated with $\theta_1$ is $[0.2 \, \ 0.9]'$ and we add the elements $0.2+0.9=1.1$. The column vector associated with $\theta_2$ is $[0.8 \, \ 0.1]'$ and we add the elements $0.8+0.1=0.9$.\\

Let $w_{12}(\theta_1)$, $w_{12}(\theta_2)$ be the parts from the conflicting mass to be assigned to $m_\cap(\theta_1)$ and $m_\cap(\theta_2)$ respectively.  Then:

$$\frac{w_{12}(\theta_1)}{1.1}=\frac{w_{12}(\theta_2)}{0.9}=\frac{w_{12}(\theta_1)+w_{12}(\theta_2)}{1.1+0.9}=\frac{0.74}{2}=0.37$$

Whence, $w_{12}(\theta_1)=1.1\cdot 0.37=0.407$, $w_{12}(\theta_2) = 0.9\cdot 0.37=0.333$. One adds $w_{12}(\theta_1)$ to $m_\cap(\theta_1)$ and $w_{12}(\theta_2)$ to $m_\cap(\theta_2)$ and one finally gets the result of the PCR1 rule of combination:

$$m_{PCR1}(\theta_1)=0.18+0.407=0.587$$
$$m_{PCR1}(\theta_2)=0.08+0.333=0.413$$

\noindent 
where $m_{PCR1}(.)$ means the normalized mass resulting from the PCR1 rule of combination.\\

We can directly use the PCR1 formula for computing the mass, instead of doing proportionalizations all the time.

\subsubsection{Fusion with the Dempster's rule}

Based on the close-world Shafer's model and applying the Dempster's rule of combination, one gets (index $DS$ standing here for Dempster-Shafer)

$$m_{DS}(\theta_1)=\frac{m_\cap(\theta_1)}{1-k_{12}}=\frac{0.18}{0.26}=0.692308$$
$$m_{DS}(\theta_2)=\frac{m_\cap(\theta_2)}{1-k_{12}}=\frac{0.08}{0.26}=0.307692$$

\subsubsection{Fusion with the Smets' rule}

Based on the open-world model with TBM interpretation \cite{Smets_1990} and applying the Smets' rule of combination (i.e. the non-normalized Dempster's rule of combination), one trivially gets (index $S$ standing here for Smets)

$$m_{S}(\theta_1)=m_\cap(\theta_1)=0.18$$
$$m_{S}(\theta_2)=m_\cap(\theta_2)=0.08$$
$$m_{S}(\emptyset)=k_{12}=0.74$$

\subsubsection{Fusion with other rules}

While different in their essence, the Yager's rule \cite{Yager_1983}, Dubois-Prade \cite{Dubois_1988}  rule and the hybrid DSm rule \cite{DSmTBook_2004a} of combination provide the same result for this specific 2D example. That is
$$m(\theta_1)= 0.18 \qquad m(\theta_2)=0.08 \qquad m(\theta_1\cup \theta_2)=0.74$$

\subsection{Example 3 (Zadeh's example)}

Let's consider the famous Zadeh's examples \cite{Zadeh_1979,Zadeh_1984,Zadeh_1985,Zadeh_1986} with the frame $\Theta=\{\theta_1,\theta_2,\theta_3\}$, two independent sources of evidence corresponding to the following Bayesian belief assignment matrix (where columns 1, 2 and 3 correspond respectively to elements $\theta_1$, $\theta_2$ and $\theta_3$ and rows 1 and 2 to belief assignments $m_1(.)$ and $m_2(.)$ respectively), i.e.

\begin{equation*}
\mathbf{M}_{12}=
\begin{bmatrix}
0.9 & 0 & 0.1\\
0 & 0.9 & 0.1
\end{bmatrix}
\end{equation*}
\noindent 
In this example, one has
$$\begin{cases}
m_\cap(\theta_1)=[m_1\oplus m_2](\theta_1)=0\\
m_\cap(\theta_2)=[m_1\oplus m_2](\theta_2)=0\\
m_\cap(\theta_3)=[m_1\oplus m_2](\theta_3)=0.1\cdot 0.1=0.01\\
\end{cases}$$
\noindent 
and the conflict between the sources is very high and is given by $$k_{12}=1 - m_\cap(\theta_1) - m_\cap(\theta_2) - m_\cap(\theta_3)= 0.99$$

\subsubsection{Fusion with the PCR1 rule}

Using the PCR1 rule of combination, the conflict $k_{12}=0.99$ is proportionally distributed to $m_\cap(\theta_1)$, $m_\cap(\theta_2)$, $m_\cap(\theta_3)$ with respect to their corresponding sums of columns, i.e. $0.9$, $0.9$, $0.2$ respectively. Thus: $w_{12}(\theta_1)/0.9 = w_{12}(\theta_2)/0.9 = w_{12}(\theta_3)/0.2 = 0.99/2 = 0.495$. Hence: $w_{12}(\theta_1)=0.9\cdot 0.495=0.4455$, $w_{12}(\theta_2) =0.9\cdot 0.495=0.4455$ and $w_{12}(\theta_3)=0.2\cdot 0.495=0.0990$. Finally the result of the PCR1 rule of combination is given by
$$m_{PCR1}(\theta_1)=0+0.4455=0.4455$$
$$m_{PCR1}(\theta_2)=0+0.4455=0.4455$$
$$m_{PCR1}(\theta_3)=0.01+0.099=0.109$$
This is an acceptable result if we don't want to introduce the partial ignorances (epistemic partial uncertainties). This result is close to Murphy's arithmetic mean combination rule \cite{Murphy_2000}, which is the following ($M$ index standing here for the Murphy's rule) :
$$m_{M}(\theta_1)=(m_1(\theta_1)+m_2(\theta_1))/2 = (0.9+0)/2 = 0.45$$
$$m_{M}(\theta_2)=(m_1(\theta_2)+m_2(\theta_2))/2 = (0+0.9)/2 = 0.45$$
$$m_{M}(\theta_3)=(m_1(\theta_3)+m_2(\theta_3))/2 = (0.1+0.1)/2 = 0.10$$

\subsubsection{Fusion with the Dempster's rule}

The use of the Dempster's rule of combination yields here to the counter-intuitive result $m_{DS}(\theta_3)=1$. This example is discussed in details in \cite{DSmTBook_2004a} where several other infinite classes of counter-examples to the Dempster's rule are also presented.

\subsubsection{Fusion with the Smets' rule}

Based on the open-world model with TBM, the Smets' rule of combination gives very little information, i;e. $m_S(\theta_3) =0.01$ and $m_S(\emptyset)=k_{12}=0.99$.

\subsubsection{Fusion with the Yager's rule}

The Yager's rule of combination transfers the conflicting mass $k_{12}$ onto the total uncertainty and thus provides little specific information since one gets $m_Y(\theta_3) =0.01$ and $m_Y(\theta_1\cup \theta_2\cup \theta_3)=0.99$.

\subsubsection{Fusion with the Dubois \& Prade and DSmT rule}

In zadeh's example, the hybrid DSm rule and the Dubois-Prade rule give the same result: $m(\theta_3)=0.01 $, $m(\theta_1\cup \theta_2)=0.81$, $m(\theta_1\cup \theta_3)=0.09$ and $m(\theta_2\cup \theta_3)=0.09$.
This fusion result is more informative/specific than previous rules of combination and is acceptable if one wants to take into account all aggregated partial epistemic uncertainties.

\subsection{Example 4 (with total conflict)}

Let's consider now the 4D case with the frame $\Theta=\{\theta_1,\theta_2,\theta_3,\theta_4\}$ and two independent equi-reliable sources of evidence with the following Bayesian belief assignment matrix (where columns 1, 2, 3 and 4 correspond to elements $\theta_1$, $\theta_2$, $\theta_3$ and $\theta_4$ and rows 1 and 2 to belief assignments $m_1(.)$ and $m_2(.)$ respectively)
\begin{equation*}
\mathbf{M}_{12}=
\begin{bmatrix}
0.3 & 0 & 0.7 & 0\\
0 & 0.4 & 0 & 0.6
\end{bmatrix}
\end{equation*}

\subsubsection{Fusion with the PCR1 rule}

Using the PCR1 rule of combination, one gets $k_{12}=1$ and
$$m_\cap(\theta_1)=m_\cap(\theta_2)=m_\cap(\theta_3)=m_\cap(\theta_4)=0$$
We distribute the conflict among $m_\cap(\theta_1)$, $m_\cap(\theta_2)$, $m_\cap(\theta_3)$ and $m_\cap(\theta_4)$ proportionally with their sum of columns, i.e., $0.3$, $0.4$, $0.7$ and $0.6$ respectively.
Thus: 
 \begin{equation*}
\frac{w_{12}(\theta_1)}{0.3}=\frac{w_{12}(\theta_2)}{0.4}=\frac{w_{12}(\theta_3)}{0.7}=\frac{w_{12}(\theta_4)}{0.6}
=\frac{w_{12}(\theta_1)+w_{12}(\theta_2)+w_{12}(\theta_3)+w_{12}(\theta_4)}{0.3+0.4+0.7+0.6}=\frac{1}{2}=0.5
 \end{equation*}

Then $w_{12}(\theta_1)=0.3\cdot 0.5=0.15$, $w_{12}(\theta_2)=0.4\cdot 0.5 = 0.20$, $w_{12}(\theta_3)=0.7\cdot 0.5=0.35$ and $w_{12}(\theta_4)=0.6\cdot 0.5=0.30$
and add them to the previous masses.  One easily gets:
$$m_{PCR1}(\theta_1)=0.15\qquad m_{PCR1}(\theta_2)=0.20\qquad m_{PCR1}(\theta_3)=0.35\qquad m_{PCR1}(\theta_4)=0.30$$
 
In this case the PCR1 combination rule gives the same result as Murphy's arithmetic mean combination rule.

\subsubsection{Fusion with the Dempster's rule}

In this example, the Dempster's rule can't be applied since the sources are in total contradiction because $k_{12}=1$. Dempster's rule is mathematically not defined because of the indeterminate form 0/0.

\subsubsection{Fusion with the Smets'  rule}

Using open-world assumption, the Smets' rule provides no specific information, only $m_S(\emptyset)=1$.

\subsubsection{Fusion with the Yager's  rule}

The Yager's rule gives no information either: $m_Y(\theta_1\cup \theta_2\cup \theta_3\cup \theta_4)=1$ (total ignorance).

\subsubsection{Fusion with the Dubois \& Prade and DSmT rule}

The hybrid DSm rule and the Dubois-Prade rule give here the same result:
$$m(\theta_1\cup \theta_2)=0.12\qquad  m(\theta_1\cup \theta_4)=0.18\qquad m(\theta_2\cup \theta_3)=0.28\qquad m(\theta_3\cup \theta_4)=0.42$$

\subsection{Example 5 (convergent to idempotence)}

Let's consider now the 2D case with the frame of discernment $\Theta=\{\theta_1,\theta_2\}$ and two independent equi-reliable sources of evidence with the following Bayesian belief assignment matrix (where columns 1 and 2 correspond  to elements $\theta_1$ and $\theta_2$ and rows 1 and 2 to belief assignments $m_1(.)$ and $m_2(.)$ respectively)
\begin{equation*}
\mathbf{M}_{12}=
\begin{bmatrix}
0.7 & 0.3 \\
0.7 & 0.3
\end{bmatrix}
\end{equation*}
The conjunctive consensus yields here:
$$m_{\cap}(\theta_1)=0.49\qquad\text{and}\qquad m_\cap(\theta_2)=0.09$$
\noindent 
with conflict $k_{12}=0.42$.

\subsubsection{Fusion with the PCR1 rule}

Using the PCR1 rule of combination, one gets after distributing the conflict proportionally among $m_{\cap}(\theta_1)$ and $m_{\cap}(\theta_2)$ with $0.7+0.7=1.4$ and $0.3+0.3=0.6$ such that
$$\frac{w_{12}(\theta_1)}{1.4}=\frac{w_{12}(\theta_2)}{0.6}=\frac{w_{12}(\theta_1)+w_{12}(\theta_2)}{1.4+0.6}=\frac{0.42}{2}=0.21$$
\noindent
whence $w_{12}(\theta_1)=0.294$ and $w_{12}(\theta_2)=0.126$ involving the following result
$$m_{PCR1}(\theta_1)=0.49+0.294=0.784\qquad m_{PCR1}(\theta_2)=0.09+0.126=0.216$$

\subsubsection{Fusion with the Dempster's rule}

The Dempster's rule of combination gives here:
$$m_{DS}(\theta_1)=0.844828\qquad \text{and}\qquad m_{DS}(\theta_2)=0.155172$$

\subsubsection{Fusion with the Smets' rule}

Based on the open-world model with TBM, the Smets' rule of combination provides here:
$$m_S(\theta_1)=0.49 \qquad m_S(\theta_2)=0.09 \qquad m_S(\emptyset)=0.42$$

\subsubsection{Fusion with the other  rules}

The hybrid DSm rule, the Dubois-Prade rule and the Yager's give here:
$$m(\theta_1)=0.49\qquad m(\theta_2)=0.09\qquad m(\theta_1\cup \theta_2)=0.42$$

\subsubsection{Behavior of the PCR1 rule with respect to idempotence}

Let's combine now with the PCR1 rule four equal sources $m_1(.)=m_2(.)=m_3(.)=m_4(.)$ with
$m_i(\theta_1)=0.7$ and $m_i(\theta_2)=0.3$ ($i=1,\ldots ,4$). The PCR1 result\footnote{The verification is left to the reader.} is now given by $$m_{PCR1}^{1234}(\theta_1)=0.76636\qquad m_{PCR1}^{1234}(\theta_2)=0.23364$$
Then repeat the fusion with the PCR1 rule for eight equal sources $m_i(\theta_1)=0.7$ and $m_i(\theta_2)=0.3$ ($i=1,\ldots ,8$). One gets now:
$$m_{PCR1}^{1\ldots 8}(\theta_1)=0.717248\qquad m_{PCR1}^{1\ldots 8}(\theta_2)=0.282752$$
Therefore $m_{PCR1}(\theta_1) \rightarrow 0.7$ and $m_{PCR1}(\theta_2) \rightarrow 0.3$. We can prove that the fusion using PCR1 rule converges towards idempotence, i.e. for $i=1,2$
$$\lim_{n\rightarrow\infty} \underbrace{[m\oplus m\oplus\ldots\oplus m](\theta_i)}_{n\,\text{times}}=m(\theta_i)$$

\noindent 
in the 2D simple case with exclusive hypotheses, no unions, neither intersections (i.e. with Bayesian belief assignments).\\

Let $\Theta=\{\theta_1,\theta_2\}$ and the mass matrix 
\begin{equation*}
\mathbf{M}_{1\ldots n}=
\begin{bmatrix}
a & 1-a \\
a & 1-a \\
\vdots & \vdots\\
a & 1-a \\
\end{bmatrix}
\end{equation*}
Using the general PCR1 formula, one gets for any $A\neq\emptyset$,
\begin{equation*}
\lim_{n\rightarrow\infty} m^{1\ldots n}_{PCR1}(\theta_1) = a^n + n\cdot a\cdot \frac{k_{1\ldots n}}{n}= a^n + a [1- a^n -{(1-a)}^n]=a
\end{equation*}
\noindent
because $\lim_{n\rightarrow\infty} a^n = \lim_{n\rightarrow\infty} {(1-a)}^n=0$ when $0<a<1$; if $a=0$ or $a=1$ also $\lim_{n\rightarrow\infty} m^{1\ldots n}_{PCR1}(\theta_1)=a$. We can prove similarly 
$\lim_{n\rightarrow\infty} m^{1\ldots n}_{PCR1}(\theta_2)=1-a$\\

One similarly proves the n-D, $n\geq 2$, simple case for $\Theta=\{\theta_1,\theta_2,\ldots,\theta_n\}$ with exclusive elements when no mass is on unions neither on intersections.

\subsection{Example 6 (majority opinion)}

Let's consider now the 2D case with the frame $\Theta=\{\theta_1,\theta_2\}$ and two independent equi-reliable sources of evidence with the following belief assignment matrix (where columns 1 and 2 correspond  to elements $\theta_1$ and $\theta_2$ and rows 1 and 2 to belief assignments $m_1(.)$ and $m_2(.)$ respectively)
\begin{equation*}
\mathbf{M}_{12}=
\begin{bmatrix}
0.8 & 0.2 \\
0.3 & 0.7
\end{bmatrix}
\end{equation*}

Then after a while, assume that a third independent source of evidence is introduces with belief assignment $m_3(\theta_1)=0.3$ and $m_3(\theta_2)=0.7$. The previous belief matrix is then extended/updated as follows (where the third row of matrix $\mathbf{M}$ corresponds to the new source $m_3(.)$)
\begin{equation*}
\mathbf{M}_{123}=
\begin{bmatrix}
0.8 & 0.2 \\
0.3 & 0.7\\
0.3 & 0.7
\end{bmatrix}
\end{equation*}

\subsubsection{Fusion with the PCR1 rule}

The conjunctive consensus for sources 1 and 2 gives (where upper index 12 denotes the fusion of source 1 and 2)
$$m^{12}_\cap(\theta_1)=0.24 \qquad  m^{12}_\cap(\theta_2)=0.14$$
\noindent 
with conflict $k_{12}=0.62$.\\

We distribute the conflict 0.62 proportionally with 1.1 and 0.9 respectively to $m^{12}_\cap(\theta_1)$ and $m^{12}_\cap(\theta_2)$ such that
$$\frac{w_{12}(\theta_1)}{1.1}=\frac{w_{12}(\theta_2)}{0.9}=\frac{w_{12}(\theta_1)+w_{12}(\theta_2)}{1.1+0.9}=\frac{0.62}{2}=0.31$$ 
\noindent 
and thus $w_{12}(\theta_1)=1.1\cdot 0.31=0.341$ and $w_{12}(\theta_2)=0.9\cdot 0.31=0.279$.\\

Using the PCR1 combination rule for sources 1 and 2, one gets:
$$m^{12}_{PCR1}(\theta_1)=0.24+0.341=0.581\qquad m^{12}_{PCR1}(\theta_2)=0.14+0.279=0.419$$
Let's combine again the previous result with $m_3(.)$ to check the majority rule (if the result's trend is towards $m_3=m_2$). Consider now the following matrix (where columns 1 and 2 correspond  to elements $\theta_1$ and $\theta_2$ and rows 1 and 2 to belief assignments $m^{12}_{PCR1}(.)$ and $m_3(.)$ respectively)
\begin{equation*}
\mathbf{M}_{12,3}=
\begin{bmatrix}
0.581 & 0.419 \\
0.3 & 0.7\\
\end{bmatrix}
\end{equation*}
The conjunctive consensus obtained from $m^{12}_{PCR1}(.)$ and $m_3(.)$  gives
$$m^{12,3}_\cap(\theta_1)=0.1743 \qquad  m^{12,3}_\cap(\theta_2)=0.2933$$
\noindent 
with conflict $k_{12,3}=0.5324$ where the index notation 12,3 stands here for the combination of the result of the fusion of sources 1 and 2 with the new source 3. The proportionality coefficients are obtained from
\begin{equation*}
\frac{w_{12}(\theta_1)}{0.581+0.3}=\frac{w_{12}(\theta_2)}{0.419+0.7}=\frac{w_{12}(\theta_1)+w_{12}(\theta_2)}{0.581+0.3+0.419+0.7}=\frac{0.5324}{2}=0.2662
\end{equation*}
\noindent 
and thus: $$w_{12}(\theta_1)=0.881\cdot 0.2662=0.234522\qquad w_{12}(\theta_2)=1.119\cdot 0.2662=0.297878$$
The fusion result obtained by the PCR1 after the aggregation of sources 1 and 2 with the new source 3 is:
$$m^{12,3}_{PCR1}(\theta_1)=0.1743+0.234522=0.408822\qquad m^{12,3}_{PCR1}(\theta_2)= 0.2933+0.297878=0.591178$$
\noindent
Thus $\mathbf{m}^{12,3}_{PCR1}=[0.408822 \, \ 0.591178]$ starts to reflect the majority opinion $\mathbf{m}_2(.)=\mathbf{m}_3=[0.3 \, \ 0.7]$ (i.e. the mass of $\theta_1$ becomes smaller than the mass of $\theta_2$).\\

If now we apply the PCR1 rule for the 3 sources taken directly together, one gets
$$m_\cap^{123}(\theta_1)=0.072 \qquad m_\cap^{123}(\theta_2)=0.098$$

\noindent 
with the total conflicting mass $k_{123}=0.83$.\\

Applying the proportionalization from $\mathbf{M}_{123}$, one has
\begin{equation*}
\frac{w_{123}(\theta_1)}{0.8+0.3+0.3}=\frac{w_{123}(\theta_2)}{0.2+0.7+0.7}=\frac{w_{123}(\theta_1)+w_{123}(\theta_2)}{3}=\frac{0.83}{3}
\end{equation*}
\noindent 
Thus, the proportionalized conflicting masses to transfer onto $m_\cap^{123}(\theta_1)$ and $m_\cap^{123}(\theta_2)$ are respectively given by
$$w_{123}(\theta_1)=1.4\cdot \frac{0.83}{3}= 0.387333\qquad w_{123}(\theta_2)=1.6\cdot \frac{0.83}{3}= 0.442667$$
The final result of the PCR1 rule combining all three sources together is then
$$m^{123}_{PCR1}(\theta_1)=0.072+0.387333=0.459333\qquad m^{123}_{PCR1}(\theta_2)= 0.098+0.442667=0.540667$$
The majority opinion is reflected since $m^{123}_{PCR1}(\theta_1) < m^{123}_{PCR1}(\theta_2)$.
Note however that the PCR1 rule of combination is clearly not associative because $(m^{12,3}_{PCR1}(\theta_1)=0.408822)\neq (m^{123}_{PCR1}(\theta_1)=0.459333)$ and $(m^{12,3}_{PCR1}(\theta_2)= 0.591178)\neq (m^{123}_{PCR1}(\theta_2)=0.540667)$.\\

If we now combine the three previous sources altogether with the fourth source providing the majority opinion, i.e. $m_4(\theta_1)=0.3$ and $m_4(\theta_2)=0.7$ one will get
$$m_\cap^{1234}(\theta_1)=0.0216 \qquad m_\cap^{123}(\theta_2)=0.0686$$
\noindent 
with the total conflicting mass $k_{1234}=0.9098$.\\

Applying the proportionalization from mass matrix
\begin{equation*}
\mathbf{M}_{1234}=
\begin{bmatrix}
0.8 & 0.2 \\
0.3 & 0.7\\
0.3 & 0.7\\
0.3 & 0.7
\end{bmatrix}
\end{equation*}
\noindent
yields 
$$w_{1234}(\theta_1)=[0.8+0.3+0.3+0.3]\cdot \frac{0.9098}{4}\qquad w_{1234}(\theta_2)=[0.2+0.7+0.7+0.7]\cdot \frac{0.9098}{4}$$
and finally the followwing result
\begin{equation*}
m^{1234}_{PCR1}(\theta_1)=0.0216+[0.8+0.3+0.3+0.3]\cdot \frac{0.9098}{4}=0.408265
\end{equation*}
\begin{equation*}
m^{1234}_{PCR1}(\theta_2)= 0.0686+[0.2+0.7+0.7+0.7]\cdot \frac{0.9098}{4}=0.591735
\end{equation*}
Hence $m^{1234}_{PCR1}(\theta_1)$ is decreasing more and more while $m^{1234}_{PCR1}(\theta_2)$ is increasing more and more, which reflects again the majority opinion.

\subsection{Example 7 (multiple sources of information)}

Let's consider now the 2D case with the frame $\Theta=\{\theta_1,\theta_2\}$ and 10 independent equi-reliable sources of evidence with the following Bayesian belief assignment matrix (where columns 1 and 2 correspond  to elements $\theta_1$ and $\theta_2$ and rows 1 to 10  to belief assignments $m_1(.)$ to $m_{10}(.)$ respectively)
\begin{equation*}
\mathbf{M}_{1\ldots10}=
\begin{bmatrix}
1 & 0 \\
0.1 & 0.9 \\
0.1 & 0.9 \\
0.1 & 0.9 \\
0.1 & 0.9 \\
0.1 & 0.9 \\
0.1 & 0.9 \\
0.1 & 0.9 \\
0.1 & 0.9 \\
0.1 & 0.9
\end{bmatrix}
\end{equation*}

\noindent
The conjunctive consensus operator gives here
$$m_\cap(\theta_1)={(0.1)}^9 \qquad m_\cap(\theta_2)=0$$
\noindent 
with the conflict $k_{1\ldots 10}=1-{(0.1)}^9$.

\subsubsection{Fusion with the PCR1 rule}

Using the general PCR1 formula \eqref{eq:PCRn}, one gets
\begin{align*}
m_{PCR1}^{1\ldots 10}(\theta_1)&={(0.1)}^9 + c_{1\ldots 10}(\theta_1)\cdot\frac{k_{1\ldots 10}}{10}
={(0.1)}^9 + (1.9)\cdot \frac{1-{(0.1)}^9}{10}
={(0.1)}^9 + (0.19)\cdot [1-{(0.1)}^9]\\
&={(0.1)}^9 + 0.19 - 0.19\cdot {(0.1)}^9={(0.1)}^9\cdot 0.81 + 0.19\approx 0.19
\end{align*}
\begin{align*}
m_{PCR1}^{1\ldots 10}(\theta_2)&={(0.9)}^9 + c_{1\ldots 10}(\theta_2)\cdot\frac{k_{1\ldots 10}}{10}
={(0.9)}^9 + (8.1)\cdot \frac{1-{(0.1)}^9}{10}
={(0.9)}^9 + (0.81)\cdot [1-{(0.1)}^9]\\
&={(0.9)}^9 + 0.81 - 0.81\cdot {(0.1)}^9={(0.1)}^9\cdot 0.19 + 0.81\approx 0.81
\end{align*}

The PCR1 rule's result is converging towards the Murphy's rule in this case, which is $m_M(\theta_1)=0.19$ and $m_M(\theta_2)=0.81$.

\subsubsection{Fusion with the Dempster's rule}

In this example, the Dempster's rule of combination gives $m_{DS}(\theta_1)=1$ which looks quite surprising and certainly wrong since nine sources indicate $m_i(\theta_1)=0.1$ ($i=2,\ldots,10$) and only one shows $m_1(\theta_1)=1$.

\subsubsection{Fusion with the Smets' rule}

In this example when assuming open-world model, the Smets' rule provide little specific information since one gets
$$m_S(\theta_1)={(0.1)}^9 \qquad m_S(\emptyset)=1 - {(0.1)}^9$$

\subsubsection{Fusion with the other  rules}

The hybrid DSm rule, the Dubois-Prade's rule and the Yager's rule give here:
$$m(\theta_1)={(0.1)}^9 \qquad m(\theta_1\cup \theta_2)=1 - {(0.1)}^9$$
\noindent
which is less specific than PCR1 result but seems more reasonable and cautious if one introduces/takes into account  epistemic uncertainty arising from the conflicting sources if we consider that the majority opinion does not necessary reflect the reality of the solution of a problem. The answer to this philosophical question is left to the reader.

\subsection{Example 8 (based on hybrid DSm model)}

In this last example, we show how the PCR1 rule can be applied on a fusion problem characterized by a hybrid DSm model rather than the Shafer's model and we compare the result of the PCR1 rule with the result obtained from the hybrid DSm rule.\\

Let's consider a 3D case (i.e. $\Theta=\{\theta_1,\theta_2,\theta_2\}$) including epistemic uncertainties with the two following belief assignments
$$m_1(\theta_1)=0.4 \qquad m_1(\theta_2)=0.1\qquad m_1(\theta_3)=0.3\qquad m_1(\theta_1\cup\theta_2)=0.2$$
$$m_2(\theta_1)=0.6\qquad m_2(\theta_2)=0.2\qquad m_2(\theta_3)=0.2$$

We assume here a {\it{hybrid DSm model}} \cite{DSmTBook_2004a} (chap. 4) in which the following integrity constraints hold

$$\theta_1\cap \theta_2=\theta_1\cap\theta_3=\emptyset$$
\noindent
but where $\theta_2\cap\theta_3\neq\emptyset$.\\

The conjunctive consensus rule extended to the hyper-power set $D^\Theta$ (i.e. the Dedekind's lattice built on $\Theta$ with union and intersection operators) becomes now the classic DSm rule and we obtain
$$m_\cap(\theta_1)=0.36\qquad m_\cap(\theta_2)=0.06\qquad m_\cap(\theta_3)=0.06\qquad  m_\cap(\theta_2\cap\theta_3)=0.12$$
\noindent
One works on hyper-power set (which contains, besides unions, intersections as well), not on power set as in all other theories based on the Shafer's model (because power set contains only unions, not intersections).\\

The conflicting mass $k_{12}$ is thus formed together by the masses of $\theta_1\cap \theta_2$ and $\theta_1\cap \theta_3$ and is given by
\begin{align*}
k_{12}&=m(\theta_1\cap \theta_2)+m(\theta_1\cap \theta_3)=[0.4\cdot 0.2 + 0.6\cdot 0.1] + [0.4\cdot 0.2 + 0.6\cdot 0.2] = 0.14 + 0.26=0.40\\
& = 1 - m_\cap(\theta_1) -m_\cap(\theta_2)- m_\cap(\theta_3)-m_\cap(\theta_2\cap\theta_3)
\end{align*}

The classic DSm rule (denoted here with index DSmc) provides also
$$m_{DSmc}(\theta_2\cap \theta_3)=0.1\cdot 0.2 + 0.2\cdot 0.3=0.08\qquad m_{DSmc}(\theta_3\cap (\theta_1\cup\theta_2))=0.04$$
\noindent 
but since $\theta_3\cap (\theta_1\cup\theta_2)=(\theta_3\cap \theta_1)\cup (\theta_3\cap \theta_2)=\theta_2\cap \theta_3$ because integrity constraint $\theta_1\cap\theta_3=\emptyset$ of the model, the total mass committed to $\theta_2\cap \theta_3$ is finally
$$m_{DSmc}(\theta_2\cap \theta_3)=0.08+0.04=0.12$$

\subsubsection{Fusion with the hybrid DSm rule}

If one uses the hybrid DSm rule, one gets
$$m_{DSmh}(\theta_1)=0.36\qquad m_{DSmh}(\theta_2)=0.06\qquad m_{DSmh}(\theta_3)=0.06$$
$$m_{DSmh}(\theta_1\cup \theta_2)=0.14\qquad m_{DSmh}(\theta_1\cup \theta_3)=0.26\qquad m_{DSmh}(\theta_2\cap \theta_3)=0.12$$

\subsubsection{Fusion with the PCR1 rule}

If one uses the PCR1 rule, one has to distribute the conflicting mass 0.40 to the others according to
\begin{equation*}
\frac{w_{12}(\theta_1)}{1.0}=\frac{w_{12}(\theta_2)}{0.3}=\frac{w_{12}(\theta_3)}{0.5}
=\frac{w_{12}(\theta_1\cup\theta_2)}{0.2}=\frac{0.40}{2}=0.20
\end{equation*}

Thus one deduces $w_{12}(\theta_1)=0.20$, $w_{12}(\theta_2)=0.06$, $w_{12}(\theta_3)=0.10$ and $w_{12}(\theta_1\cup\theta_2)=0.04$.\\

Nothing is distributed to $\theta_1\cup\theta_2$ because its column in the mass matrix is $[0\, \ 0]'$, therefore its sum is zero. Finally, one gets the following results with the PCR1 rule of combination:
$$m_{PCR1}(\theta_1)=0.36+0.20=0.56\qquad m_{PCR1}(\theta_2)=0.06+0.06=0.12\qquad m_{PCR1}(\theta_3)=0.06+0.10=0.16$$
$$m_{PCR1}(\theta_1\cup\theta_2)=0+0.0.4=0.04\qquad m_{PCR1}(\theta_2\cap\theta_3)=0.12+0=0.12$$

\section{ Conclusion}

In this paper a very simple alternative rule to WAO has been proposed for managing the transfer of epistemic uncertainty in any framework (Dempster-Shafer Theory, Dezert-Smarandache Theory) which overcomes limitations of the Dempster's rule yielding to counter-intuitive results for highly conflicting sources to combine. This rule is interesting both from the implementation standpoint and the coherence of the result if we don't accept the transfer of conflicting mass to partial ignorances. It appears as an interesting compromise between the Dempster's rule of combination and the more complex (but more cautious) hybrid DSm rule of combination. This first and simple Proportional Conflict Redistribution (PCR1) rule of combination works in all cases no matter how big the conflict is between sources, but when some sources become totally ignorant because in such cases, PCR1 (as WAO) does not preserve the neutrality property of the vacuous belief assignment in the combination. PCR1 corresponds to a given choice of proportionality coefficients in the infinite continuum family of possible rules of combination (i.e.  weighted operator - WO) involving conjunctive consensus pointed out by Inagaki in 1991 and Lef\`evre, Colot and Vannoorenberghe in 2002. The PCR1 on the power set and for non-degenerate cases gives the same results as WAO; yet, for the storage proposal in a dynamic fusion when the associativity is needed, for PCR1 it is needed to store only the last sum of masses, besides the previous conjunctive rules result, while in WAO it is in addition needed to store the number of the steps. PCR1 and WAO rules become quasi-associative. In this work, we extend WAO (which herein gives no solution) for the degenerate case when all column sums of all non-empty sets are zero, and then the conflicting mass is transferred to the non-empty disjunctive form of all non-empty sets together; but if this disjunctive form happens to be empty, then one considers an open world (i.e. the frame of discernment might contain new hypotheses) and thus all conflicting mass is transferred to the empty set. In addition to WAO, we propose a general formula for PCR1 (WAO for non-degenerate cases).  Several numerical examples and comparisons with other rules for combination of evidence published in literature have been presented too. Another distinction between these alternative rules is that WAO is defined on the power set $2^\Theta$, while PCR1 is on the hyper-power set $D^\Theta$. PCR1 and WAO are particular cases of the WO. In PCR1, the proportionalization is done for each non-empty set with respect to the non-zero sum of its corresponding mass matrix - instead of its mass column average as in WAO, but the results are the same as Ph. Smets has pointed out in non degenerate cases.  In this paper,  one has also proved that a nice feature of PCR1, is that it  works in all cases; i.e. not only on non-degenerate cases but also on degenerate cases as well (degenerate cases might appear in dynamic fusion problems), while the WAO does not work in these cases since it gives the sum of masses less than 1. WAO and PCR1 provide both however a counter-intuitive result when one or several sources become totally ignorant that why improved versions of PCR1 have been developed in a companion paper.

\section{ Acknowledgement}

We want to thank Dr. Wu Li from NASA Langley Research Center and Dr. Philippe Smets from the Universit\'e Libre de Bruxelles for their comments, corrections, and advises.

\end{document}